\documentclass{article}
\usepackage{arxiv}
\pdfoutput=1

\usepackage[utf8]{inputenc} % allow utf-8 input
\usepackage[T1]{fontenc}    % use 8-bit T1 fonts
\usepackage[skip=0ex]{caption}
\usepackage{soul, color}
\usepackage{amssymb}
\usepackage{amsmath}
\usepackage{booktabs}
\usepackage{graphicx}
\usepackage{longtable}
\usepackage{multirow}
\usepackage{multicol}
\usepackage{nomencl}
\makenomenclature
\usepackage{authblk}

\usepackage{xurl}
\usepackage[breaklinks=true]{hyperref}
\hypersetup{
    colorlinks=true,
    allcolors=blue,
    pdftitle={Near-term Solar Irradiance Forecasting under Data Transmission Constraints},
    pdfauthor={Joshua E.~Hammond, Ricardo A.~Lara Orozco, Michael Baldea, Brian A.~Korgel},
    }

\usepackage{etoolbox}
\renewcommand\nomgroup[1]{%
\item[\bfseries%
\ifstrequal{#1}{A}{General Abbreviations}{%
\ifstrequal{#1}{B}{Time Representations}{%
\ifstrequal{#1}{C}{Irradiance Representations}{}}}%
]}
\renewcommand{\nompreamble}{\begin{multicols}{2}}
\renewcommand{\nompostamble}{\end{multicols}}

\usepackage{cleveref}

\title{Short-Term Solar Irradiance Forecasting under Data Transmission Constraints}

\date{March 19, 2024}

\author[1]{%
	Joshua E.~Hammond \thanks{\texttt{joshua.hammond@utexas.edu}}}%
\author[2]{%
	Ricardo A.~Lara Orozco\thanks{\texttt{rlara@che.utexas.edu}}}%
\author[1,3]{%
    Michael Baldea\thanks{\texttt{mbaldea@che.utexas.edu}}}%
\author[1,4]{%
    Brian A. Korgel\thanks{\texttt{korgel@che.utexas.edu}}}%
\affil[1]{McKetta Department of Chemical Engineering, The University of Texas at Austin, Austin, TX 78712}
\affil[2]{Hildebrand Austin Department of Petroleum and Geosystems Engineering, The University of Texas at Austin, Austin, TX 78712}
\affil[4]{Energy Institute, The University of Texas at Austin, Austin, TX 78712}
\affil[3]{Institute for Computational Engineering and Sciences, The University of Texas at Austin, Austin, TX 78712}

\begin{document}

\maketitle

% define macros
\newcommand{\degrees}{$\!\!$\char23$\!$}
\newcommand{\needsource}{\hl{[source?]}}
\renewcommand{\UrlBreaks}{\do\/\do\%\do-\do_}
\newcommand{\wmm}{$W/m^2$}

\begin{abstract}
We report a data-parsimonious machine learning model for short-term forecasting of solar irradiance. The model inputs include sky camera images that are reduced to scalar features to meet data transmission constraints. The output irradiance values are transformed to focus on unknown short-term dynamics. Inspired by control theory, a noise input is used to reflect unmeasured variables and is shown to improve model predictions, often considerably. Five years of data from the NREL Solar Radiation Research Laboratory were used to create three rolling train-validate sets and determine the best representations for time, the optimal span of input measurements, and the most impactful model input data (features). For the chosen test data, the model achieves a mean absolute error of  74.34 $W/m^2$ compared to a baseline 134.35 $W/m^2$ using the persistence of cloudiness model.

\end{abstract}

% TODO Don't forget about Graphical Abstract



% \begin{keyword}
% %% To ask: should the keywords be PACS or MSC?
% Feature Engineering \sep
% Forecasting \sep
% Machine Learning \sep
% Renewable Energy \sep
% Solar Irradiance

% \end{keyword}

% \end{frontmatter}

\section{Introduction}
\label{sec:intro}

The United States has set an aggressive target to achieve a power grid with net-zero greenhouse gas emissions by 2035 \cite{magnan2021, guterres2022, biden2021}. This will require a major shift in the power generation mix from fossil fuels, to include significantly more renewable sources, such wind and solar \cite{biden2021, denholm2022, EIA2022}. Since wind and solar power generation are intermittent and largely non-dispatchable, it will become increasingly important to anticipate fluctuations in renewable power generation to ensure the stability of the grid. Accurate near-term forecasts of solar photovoltaic (PV) power generation will be especially important to ensure that power supply meets demand.
\nomenclature{PV}{Photovoltaic}

The economic impact of PV forecasts is considerable\cite{gandhi2024}. Accurate near-term forecasts enhance grid stability by anticipating the need for ramping events\cite{chen2020,spyrou2020,lappalainen2020,Chen2022}, and could be used to define targets for demand response and help firm generation\cite{rowe2018,bone2018}, advise real-time market price predictions\cite{cheng2020}, and inform ancillary service dispatch\cite{Li2021, Li2022}.

Short-term forecasts of PV generation rely on large volumes of diverse, high-dimensional data including local meteorological measurements, numerical weather prediction\cite{ERCOT2022}, and satellite images\cite{Paletta2022}. Sky cameras are becoming an increasingly widespread method for gathering local information at generation sites \cite{feng2020, feng2022, leguen2020}. Because of the computational power needed to make PV generation forecasts, local measurements are usually processed at a centralized facility or data center\cite{ERCOT2022}. Solar PV facilities are typically placed in remote locations, and transmitting the required data (particularly, sky camera images) from the generation sites to centralized computing facilities can be difficult due to data infrastructure limitations and costs.

It is precisely this problem that we address in this work. We introduce a  machine learning model to forecast solar irradiance over the near term, that is data parsimonious and minimizes data transmission requirements. The model uses a set meteorological station data and \emph{scalar} data extracted from sky camera images as inputs. Feature importance studies are used to discern the most impactful features in the available set. Inspired by control theory, a noise signal (along with the corresponding noise model) is introduced to capture unmeasured disturbances. The residual (difference) between the persistence of cloudiness (POC) prediction and true irradiance over the forecast horizon is utilized as the output. While the final model is less accurate than recently published sky camera-based forecasting models, it requires orders of magnitude less data transmission, and significantly outperforms reference forecasts.

The specific contributions of this work are:
\begin{itemize}
    \item A parsimonious machine learning model for near-term forecast of solar irradiance using scalar features. This model is designed to minimize data transmission requirements from remote generation sites and requires significantly less bandwidth than existing models.
    \item A novel approach to forecasting solar irradiance by predicting the deviation from the persistence of cloudiness model. This approach is predicated on eliminating the need to capture known long-term dynamics, and is shown to improve forecast accuracy (compared to including persistence information as a model input).
    \item A noise model is used -- to our knowledge, for the first time in this application area -- to account for unmeasured variables/disturbances, and is shown to further improve forecasting accuracy.
    \item New empirical insights on feature importance and the effect of input sequence length on model performance are drawn from a large-scale data set.
\end{itemize}
\nomenclature{POC}{Persistence of Cloudiness}

\printnomenclature

\section{Background}

The term ``solar forecasting'' describes the prediction of PV power generation, as well as the prediction of irradiance, given that PV generation is a function of global horizontal irradiance (GHI, $W/m^2$) and panel temperature~\cite{dobos2014}. While cloud cover and position have the largest impact on irradiance~\cite{sun2018,Sun2019,Clauzel2024} --sometimes accounting for a change in irradiance of over 80\% in a minute-- other variables such as dispersed particles and wind\cite{Bett2016} are linked to irradiance as well. As shown in \Cref{eq:ghi}, GHI is composed of both direct and diffuse sunlight, where DNI is the direct normal irradiance, $\alpha$ is the solar zenith angle, and DHI is the diffuse horizontal irradiance. Clear sky models calculate GHI in the absence of atmospheric effects using time and global position\cite{dobos2014, ineichen2008, stein2012, mikofski2017}. The ratio of measured irradiance ($GHI_t$) to ideal clear sky irradiance ($GHI_{CS,t}$) at a given time instant $t$ is often referred to as clear sky index (CSI) and is shown in \Cref{eq:csi}.

\nomenclature[C]{GHI}{Global Horizontal Irradiance}
\nomenclature[C]{DNI}{Direct Normal Irradiance}
\nomenclature[C]{DHI}{Diffuse Horizontal Irradiance}
\nomenclature[C]{GHI\(_0\)}{Irradiance at time of forecast}
\nomenclature[C]{GHI\(_t\)}{ Irradiance at time \(t\)}
% \nomenclature[C]{\(\widehat{GHI}_t\)}{Predicted Irradiance at time of forecast}
\nomenclature[C]{GHI\(_{CS,t}\)}{Ideal Clear Sky Irradiance at time \(t\)}
\nomenclature[C]{CSI\(_t\)}{Clear Sky Index at time \(t\)}

\begin{equation} \label{eq:ghi}
    GHI=\cos(\alpha)DNI + DHI
\end{equation}

\begin{equation} \label{eq:csi}
    CSI_t = \frac{GHI_t}{GHI_{CS,t}}
\end{equation}

Persistence models are often used to evaluate the effectiveness of forecasting algorithms. The persistence of cloudiness (POC) model assumes that the clear sky index at the time of forecast will remain constant  throughout the forecast horizon.  Equation \ref{eq:poc} shows an example POC forecast at 10-minute intervals up to a maximum forecast horizon of 120 minutes.

\begin{equation} \label{eq:poc}
    CSI_t= CSI_0, t \in \{10, 20, \cdots 110, 120 \}
\end{equation}

The POC  forecast does not take into account fluctuations in weather conditions, but does provide a baseline for near-term solar output forecasting. More sophisticated forecast models should achieve (significantly) higher accuracy compared to the POC model by anticipating changes in weather conditions.

It is common for these latter models to predict the \emph{difference} between the time-varying irradiance and the baseline. This has the effect of de-trending the data~\cite{hyndman2015, hyndman2021}. For example, \cite{Yang2024,Wen2023} use autoregressive models to predict irradiance relative to a clear sky model (rather than using the POC as a baseline). We note that other de-trending methods, such as wavelet decomposition~\cite{Huang2021} and  spectral decomposition~\cite{Martins2023} have been recently proposed for irradiance forecasting.

The aforementioned works are restricted to using \emph{scalar} data as model inputs,  and the number of inputs is typically small in practical settings.

There is nevertheless a wealth of data that can be relevant to irradiance forecasting. These include local meteorological measurements (cloud height, wind speed and direction, temperature, humidity, and air pressure), sky camera images, Numerical Weather Prediction (NWP) results, and satellite imaging. Some of these data are in non-scalar formats (e.g., camera images) and thus do not lend themselves naturally to use in time-series models.

\nomenclature{NWP}{Numerical Weather Prediction}

This fact has motivated the use of machine learning (ML) approaches for irradiance forecasting. Convolutional Neural Networks (CNNs) and Long Short-Term Memory (LSTM) networks are common deep learning architectures used to extract patterns from data and predict sequences, respectively\cite{chollet2021}.
\nomenclature{CNN}{Convolutional Neural Network}
\nomenclature{LSTM}{Long Short-Term Memory}

ML models broadly fall into two categories:
\begin{itemize}
\item End-to-end models utilize available data as inputs with little or no pre-processing or interpretation. For example,  \cite{peng2015} reported the use of multiple sky cameras to track cloud positions in the sky, and a stereoscopic arrangement was reported by \cite{nouri2019} for estimating cloud height. \cite{song2022} determined current irradiance directly from sky camera images, while \cite{lin2023} used deep learning models to forecast irradiance from similar data.

    Some end-to-end models are used for multiple predictions, e.g., both cloud patterns~\cite{michael2022, paletta2021a, paletta2021b} and  irradiance~\cite{feng2022, liu2021, jalali2021}. ~\cite{Liu2023} use a deep learning model to determine cloud movement vectors from image sequences as well as forecast irradiance.

    These models typically use heterogeneous data formats as inputs. It is worth recalling here the work of ~\cite{Ogliari2024}, who performed feature fusion by  leveraging unused image pixels to embed other sensor data before feeding sky-camera images to a deep CNN model.

\item Physically-motivated models take a physics-based lens to the data, generating additional insights prior to building the irradiance forecasting model itself. These include detection of cloud position, cloud motion, and identifying position of the Sun, which can be approached via color analysis \cite{ghonima2012,kazantzidis2012}, pixel clustering \cite{heinle2010,Haider2022}, and convolution methods~\cite{makwana2022}. Recent work by \cite{Paletta2022} demonstrated that polar transformations centered on the sun can extract cloud movement toward the sun. \cite{Fabel2023} use features extracted from a sky camera such as cloud coverage, height, and type as well as parametric relationships such as solar position to predict future irradiance.

\end{itemize}

We emphasize here that these are \emph{broad} categories and modeling can take inspiration from both. An example is the work of ~\cite{leguen2020}, where  physics-based fluid-flow equations to predict future irradiance using sky camera images as inputs in an end-to-end structure.

It is also of note that the works referenced here use sophisticated, computationally-intensive and data-intensive architectures to perform the irradiance forecasting task. Local data collected from -- typically remote -- solar PV generation sites are transmitted over some communication channel to a centralized location for processing and forecasting. This implicitly assumes that there are no data transmission constraints. In practice, however, the communication channel may be (severely) bandwidth limited, and may not be able to accomplish the timely transmission of all the data, particularly sky camera images which are stored in typically large files. This emphasizes the need for developing \emph{data parsimonious} models, that are capable of performing the forecasting task based on a \emph{limited} number of inputs that are readily obtainable over existing bandwidth restricted communication channels.

\section{Model}

Motivated by the above, we present a CNN-LSTM model  which uses \emph{scalar} features extracted from a commercially available sky camera to forecast near-term irradiance, expressed as the deviation of actual irradiance from the POC value.

\begin{figure}
    \centering
    \includegraphics[width=3.45in]{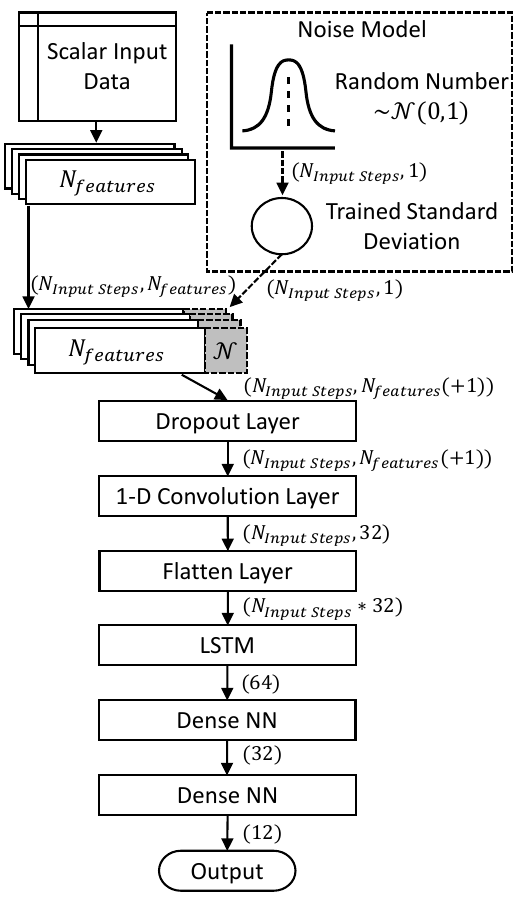}
    \caption{CNN-LSTM Model Structure with an optional noise model where random numbers drawn from a Gaussian distribution to account for unmeasured variables and disturbances. Dimensions are shown in parentheses.}
    \label{fig:architecture}
\end{figure}

\begin{figure*}
    \centering
    \includegraphics[width=7in]{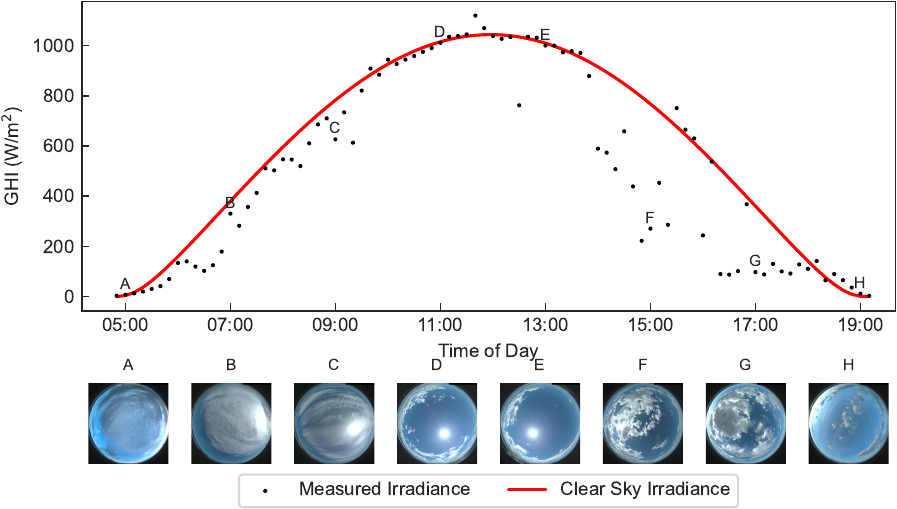}
    \caption{Irradiance measurements and a subset of sky-camera images from May 18, 2022 at the NREL SRRL BMS. Figure created by the authors based on data collected from \cite{stoffel1981}. Letters denote the moments that the images were collected.}
    \label{fig:example_day}
\end{figure*}

 The model uses up to 168 features summarized in Appendix A across five years (2017---2022) from the National Renewable Energy Laboratory (NREL) Solar Radiation Research Laboratory (SRRL) Baseline Measurement System (BMS)~\cite{stoffel1981} in Golden, Colorado, USA. The BMS provides 131 parameters including scalar data extracted from an Eko All Sky Imager (ASI-16) such as light and heavy cloud coverage percentages and whether the sun is covered by clouds~\cite{ghonima2012, EKO}. BMS data were augmented with ideal clear sky irradiance data~\cite{ineichen2008, stein2012, mikofski2017, holmgren2018} and other engineered features such as lagged statistics. \Cref{fig:example_day} shows an example of measured irradiance values compared to clear sky irradiance values, as well as selected sky images at the time of measurement. A full list of the scalar features extracted from sky camera images as well as a comprehensive list of all features examined in this work is provided in~\ref{sec:apendix-parameters}. We note that the scalar features of the sky camera images are provided by the camera itself and do not require any (additional) local processing or computing power.
\nomenclature{NREL}{National Renewable Energy Laboratory}
\nomenclature{SRRL}{Solar Radiation Research Laboratory}
\nomenclature{BMS}{Baseline Measurement System}
\nomenclature{ASI}{All Sky Imager}

While we focus on predicting irradiance at ten minute intervals up to two hours from the time of forecast, we believe that the model can be easily adapted to other forecast frequencies and prediction horizons.  % feature engineering, predicted irradiance representation, input horizon length,

\subsection{Model Architecture and Training}

The model architecture is based on using a CNN-LSTM structure as shown in Figure \autoref{fig:architecture}. A dropout layer ensures that the model learns sparse and generalizable feature representations and helps prevent overfitting. Next, CNNs efficiently extract intermediate features from input data using convolutional filters learned during model training. The LSTM layer identifies sequential patterns within the training data and is used to identify temporal patterns within the data. The final dense layers perform a bottleneck operation and transform the intermediate representations to the final irradiance sequence  prediction over the time horizon considered. We use two dense layers to ensure that the model has sufficient capacity to encode the predicted irradiance.

Similar architectures have been successfully used for irradiance prediction by e.g., \cite{gao2022} and \cite{paletta2021a}. Nevertheless, our proposed architecture include a novel feature, that, to our knowledge, has not been employed in the field of solar irradiance forecasting. Inspired by forecasting mechanisms used in the literature on model predictive control \cite{rawlings2022}, a noise signal is included as an additional input. This is motivated by the fact that it is natural to expect that the available features/data do not/cannot fully explain or predict the variable of interest (i.e., irradiance). We \emph{hypothesize} that changes in irradiance are also subject to \emph{random, unmeasured} inputs. It is typical to use a Gaussian noise signal with zero mean; additionally we assume that the variance of this signal is unity (which is a deviation from the control literature), and allow for the weight assigned to this signal to change in the training process.

The model was trained using a rolling train-validate data split with three successive steps to refine the initial model. The first step evaluated different representations of time and  irradiance. The second step studied the effect of changing the time span of input measurements on validation error. The third step used the results from the previous two steps and implemented permutation feature importance~\cite{altmann2010} to identify the most impactful features for predictive performance. The training and validation data splits for each step are shown in \Cref{tab:steps}.

\begin{table*}[ht!]
    \centering
    \caption{Model training and tuning used three successive steps to refine the initial model with an input block of 131 features and 13 temporal measurements. Count refers to the number of complete input and output data blocks within each date range.}
    \label{tab:steps}
    \begin{tabular}{llclcl}
    \toprule
    Step & Training Dates & Count & Test Dates & Count & Description \\
    \midrule
    1 & Sep. 27, 2017 --- Sep. 26, 2019 & 21,343 & Sep. 27, 2019 --- Sep. 26, 2020 & 12,509 & Time and Irradiance Representations \\
    2 & Sep. 27, 2017 --- Sep. 26, 2020 & 33,852 & Sep. 27, 2020 --- Sep. 26, 2021 & 13,256 & Input time horizon \\
    3 & Sep. 27, 2017 --- Sep. 26, 2021 & 47,108 & Sep. 27, 2021 --- Sep. 26, 2022 & 9,199 & Feature importance \\
    \bottomrule
    \end{tabular}
\end{table*}

\subsubsection{Step 1 --- Time and Irradiance Representations}

Time representations include a floating point number between 0 and 1 representing the proportion of time that has passed since the start of a day or year. These are shown as the Time of Day (ToD) and Time of Year (ToY) representations in \Cref{eq:ToD} and \Cref{eq:ToY}. Time of day may also be represented relative to key moments in the day such as sunrise, solar noon, and sunset as shown in \Cref{eq:TM}. We refer to a vector containing these three values as Time Milestones (TM). Finally, we include a trigonometric transform of ToD, ToY, and TM as is common for cyclic variables \cite{stolwijk1999}. We denote these representations as $\angle \mathrm{ToY}$, $\angle \mathrm{TM}$, and $\angle \mathrm{ToD}$ and their respective equations are shown in \Cref{eq:TrigToD}, \Cref{eq:TrigToY}, and \Cref{eq:TrigTM}.
\nomenclature[B]{ToD}{Time of Day}
\nomenclature[B]{ToY}{Time of Year}
\nomenclature[B]{TM}{Time Milestones}
\nomenclature[B]{\(\angle \)}{Sine and Cosine Transformation}

\begin{equation}
    \label{eq:ToD}
    \mathrm{ToD}=(\mathrm{Hours} + \mathrm{Minutes}/60 + \mathrm{Seconds}/3600)/24
\end{equation}

\begin{equation}
    \label{eq:ToY}
    \mathrm{ToY}=(\textnormal{Day of Year} + \mathrm{ToD})/365
\end{equation}

\begin{equation}
    \label{eq:TM}
    % \overrightarrow{\mathrm{TM}}=
    \mathrm{TM}=
    \left\{
    \begin{array}{l}
        \mathrm{ToD}(\textnormal{Time} - \textnormal{Sunrise Time}),\\
        \mathrm{ToD}(\textnormal{Time} - \textnormal{Solar Noon Time}),\\
        \mathrm{ToD}(\textnormal{Time} - \textnormal{Sunset Time})
    \end{array}
    \right\}
\end{equation}

\begin{equation}
    \label{eq:TrigToD}
    \angle \mathrm{ToD}=\{\sin(\mathrm{ToD}), \cos(\mathrm{ToD}) \}
\end{equation}

\begin{equation}
    \label{eq:TrigToY}
    \angle \mathrm{ToY}=\{\sin(\mathrm{ToY}), \cos(\mathrm{ToY}) \}
\end{equation}

\begin{equation}
    \label{eq:TrigTM}
    \angle \mathrm{TM}=\{\sin(\mathrm{TM}), \cos(\mathrm{TM}) \}
\end{equation}

Multiple representations of future irradiance were tested as target variables to determine which produced the most accurate forecast. These representations include $GHI_t$, $CSI$, CS Dev. (the difference between $GHI_t$ and $GHI_{CS}$ as shown in \Cref{eq:CSdev}), as well as two forms that are relative to the conditions at the time of forecast: change in irradiance ($\Delta \ GHI$), and change in clear sky index ($\Delta \ CSI$). These relative representations are shown in \Cref{eq:deltaGHI} and \Cref{eq:deltaCSI} respectively. Differencing and auto-regressive approaches similar to these relative representations of irradiance have been shown to improve the predictive ability of statistical models~\cite{hyndman2015, hyndman2021}.
\nomenclature[C]{CS Dev.}{Clear Sky Deviation}
\nomenclature[C]{\(\Delta GHI\)}{Change in Irradiance}
\nomenclature[C]{\(\Delta CSI\)}{Change in Clear Sky Index}

\begin{equation}
    \label{eq:CSdev}
    \mathrm{CS Dev.} = GHI_{CS} - GHI_t
\end{equation}

\begin{equation}
    \label{eq:deltaGHI}
    \Delta GHI = GHI_{t} - GHI_0
\end{equation}

\begin{equation}
    \label{eq:deltaCSI}
    \Delta CSI = CSI_t - CSI_0
\end{equation}

\subsection{Step 2 --- Input Sequence Length}

LSTM models use sequential inputs and determine temporal patterns in those inputs to predict future values. During Step 2, we investigate how the number of input measurements affects model performance.

\subsubsection{Step 3 --- Feature Importance and Noise Model Evaluation}

To determine the most important features for the model, we used permutation feature importance~\cite{altmann2010} which measures the change in model performance when a feature is randomly permuted.  To determine the effect of the noise model, we tested  model performance across all three training-validation data set splits with and without the noise model. %The model then trains the noise model variance parameter as another parameter of the model.

\subsubsection{Evaluation Metrics}

During the training process, accuracy is measured in terms of Mean Absolute Error (MAE), shown in \Cref{eq:mae}. Here, $y_i$ represents the true value and $\widehat{y}_i$ represents the predicted value. For an equitable comparison, model statistics are calculated in terms of $GHI$. To compare to other models in the literature we also calculate Root-Mean Square Error (RMSE), normalized Mean Absolute Percentage error (nMAP), as well as Forecast Skill Score (FSS) which measures performance relative to the POC model. These are defined as follows in \Cref{eq:rmse}, (\ref{eq:nmap}) and (\ref{eq:fs}) respectively:
\nomenclature[C]{\(y_i\)}{A true value}
\nomenclature[C]{\(\widehat{y}_i\)}{A predicted value}
% \nomenclature{CS Dev.}{Clear Sky Deviation}
\nomenclature{MAE}{Mean Absolute Error}
\nomenclature{RMSE}{Root Mean Square Error}
\nomenclature{nMAP}{Normalized Mean Absolute Percentage Error}
\nomenclature{FSS}{Forecast Skill Score}

\begin{equation}\label{eq:mae}
	M A E = \frac{1}{N} \sum_{i=1}^N \left|y_i-\widehat{y}_i\right|
\end{equation}

\begin{equation}\label{eq:rmse}
	R M S E =\frac{1}{N} \sum_{i=1}^N \sqrt{\left(y_i-\widehat{y}_i\right)^2}
\end{equation}

\begin{equation}\label{eq:nmap}
	n M A P=\frac{1}{N} \sum_{i=1}^N \frac{\left|y_i-\widehat{y}_i\right|}{\frac{1}{N} \sum_{i=1}^N y_i} \times 100
\end{equation}

\begin{equation}\label{eq:fs}
	FSS = 1 - \frac{y}{y_{POC}}
\end{equation}

\section{Results}

\subsection{Step 1 --- Time and Irradiance Representations}

\begin{figure}
    \centering
    \includegraphics[width=3.5in]{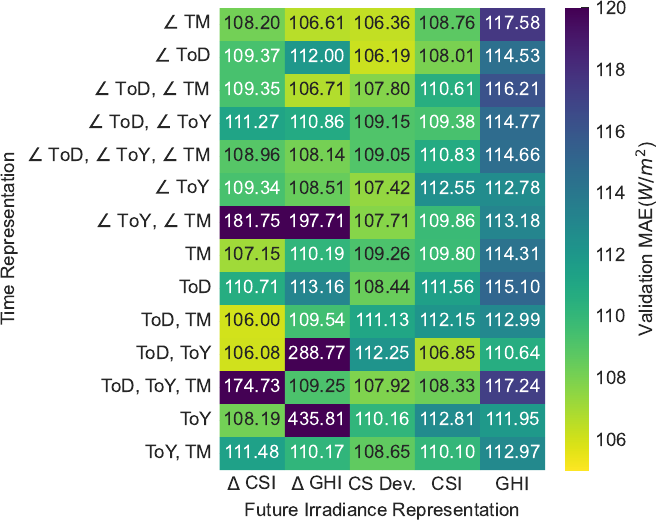}
    \caption{Validation Mean Absolute Error (W/m$^2$) across all predicted points ($t+10$ to $t+120$) for each model trained during Step 1 using year 3 as the test set. In this graphic, each row represents a different combination for time representation and each column represents a method for irradiance representation.}
    \label{fig:cv1_pivot}
\end{figure}

\autoref{fig:cv1_pivot} shows the validation MAE of all predictions (t+10 min. to t+120 min) of the model trained with different representations of time and future GHI during the first model refining step. The ToD and ToY representation of time and the $\Delta CSI$ representation of future irradiance achieve the lowest MAE across all predicted values in Step 1. Other time representations performed similarly, however, ToD and ToY allow the model to capture daily and yearly cyclical patterns without increasing the number of variables and increasing the likelihood of overfitting.

\begin{figure}
    \centering
    \includegraphics[width=3.5in]{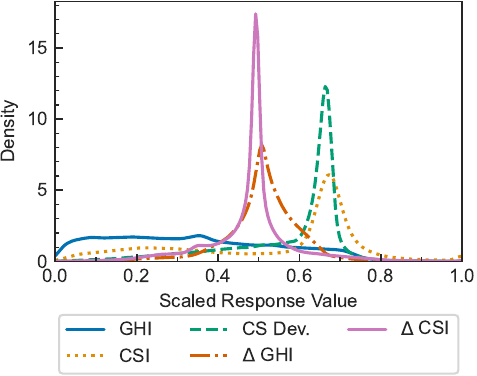}
    \caption{Probability density functions of the future irradiance representations. Note that the relative irradiance representations are much more tightly distributed around the center likely facilitating easier recognition of patterns which deviate from the present conditions.}
    \label{fig:target_distribution}
\end{figure}

$\Delta CSI$ in conjunction with ToD and ToY as time representations achieves the lowest error among the models tested in Step 1. All models trained with target variables that incorporate information from the clear sky model outperform those that use GHI. Incorporating the known patterns of the Sun using the clear sky model allows the model to isolate the unknown interest: the measured irradiance. Both relative measures of irradiance appear to be more tightly distributed around the center, as shown in \autoref{fig:target_distribution}. This likely facilitates easier recognition of patterns which deviate from the present conditions.

\begin{figure}
    \centering
    \includegraphics[width=3.5in]{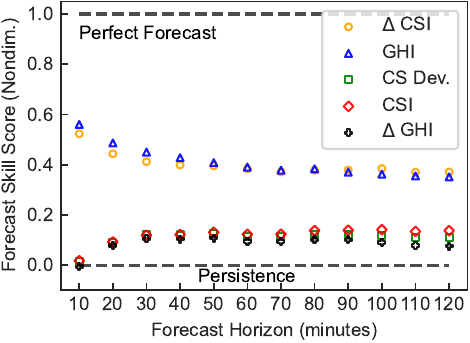}
    \caption{FSS for a model trained on each future irradiance representation.}
    \label{fig:forecastskill}
\end{figure}

\begin{figure}
    \centering
    \includegraphics[width=3.5in]{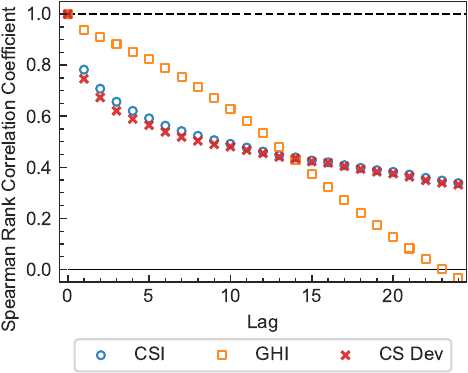}
    \caption{Autocorrelation for three representations of future irradiance beginning 2 hours or 12 lags before a prediction and extending over a two hour prediction horizon.}
    \label{fig:autocorrelation}
\end{figure}

\autoref{fig:forecastskill} shows the FSS of the best model for each future irradiance representation and demonstrates that models trained with relative representations of future irradiance are better able to anticipate changes in irradiance than models trained with other forms of irradiance. Of the two relative irradiance forms, \autoref{fig:autocorrelation} suggests that $\Delta CSI$ is likely a better target variable since clear sky index is more strongly autocorrelated (as defined in \Cref{eq:spearman}) at later lags than GHI or CS Dev.

\begin{equation}
  r_s=\frac{{cov}(\mathrm{R}(X), \mathrm{R}(Y))}{\sigma_{\mathrm{R}(X)} \sigma_{\mathrm{R}(Y)}}
\label{eq:spearman}
\end{equation}

\subsection{Step 2 --- Input Sequence Length}

\begin{figure}
    \centering
    \includegraphics[width=3.5in]{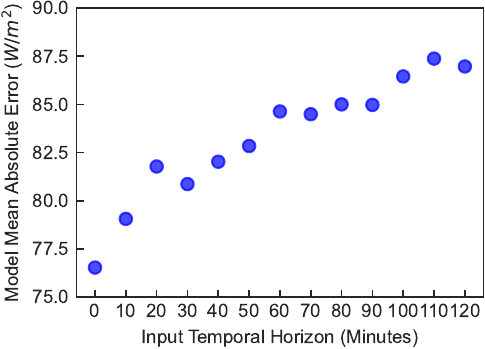}
    \caption{Mean Absolute Error (W/m$^2$) across all predicted points ($t+10$ to $t+120$) of models trained during Step 2 with increasing input time horizons. Each of these models used $\Delta$ CSI as the irradiance representation and ToD, ToY as the time representation based on the results in Step 1.}
    \label{fig:cv2_scatter}
\end{figure}

The results from varying the number of sequential measurements in Step 2 are displayed in \autoref{fig:cv2_scatter}. Surprisingly, models trained with fewer input data measurements have lower error than models that include more contextual information. However, \cite{feng2022} note that a model performs best with only the two most recent images as input. By providing less input data, the stochastic training process is less likely to result in overfitting. Because of these results, Step 3 focused on modeling an input sequence of one measurement --- an input temporal horizon of 0 minutes.

\subsection{Step 3 --- Feature Importance and Noise Model}

\begin{figure}
    \centering
    \includegraphics[width=3.5in]{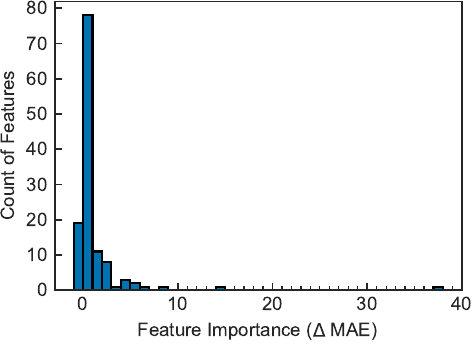}
    \caption{Feature Importance distribution for the features studied}
    \label{fig:feature_hist}
\end{figure}

\autoref{fig:feature_hist} shows the results from Step 3, a feature importance test \cite{altmann2010}, which was used to determine which of the 168 features were most meaningful for the accuracy of GHI predictions. The test systematically corrupts one feature of the input data at a time, and observes the impact on model performance. Large increases in error indicate that a particular feature is valuable to the forecast accuracy while negligible changes in error indicate that the model may not use that feature.

The ten most important features identified in Step 3 are shown in \autoref{tab:feature_top10} with the associated increase in MAE when they are corrupted. Most of the identified features (with exception to the photometers which measure dispersed aerosols) can be obtained with historical measurements, the clear sky model, and a low-cost sky-camera. Notably, two photometer measurements of Aerosols are within the top 10 results. Expensive equipment such as a photometer is not likely to be widely used in practice. When the model is retrained with the 8 remaining inputs, it achieves a MAE of 77.36 $W/m^2$, a sacrifice of only 2.16 $W/m^2$. This observation could influence equipment selection at sites interested in irradiance forecasting.

\begin{table}
\centering
\caption{Ten most important features of 131 total features as shown by the increase in predicted irradiance MAE as each feature is corrupted. Total cloud cover and GHI clear sky index contribute to prediction accuracy significantly more than other features.}
\label{tab:feature_top10}
\begin{tabular}{lr}
\toprule
Feature & Feature Importance (MAE) \\
\midrule
CDOC Total Cloud Cover & 37.76 \\
CSI GHI & 14.22 \\
DNI$_{t-4}$ & 8.08 \\
940nm Aerosols & 6.00 \\
Mean CSI DNI Deviation & 5.62 \\
675nm Aerosols & 5.20 \\
CSI DNI & 4.88 \\
Solar Elevation & 4.73 \\
CSI DNI Deviation & 4.08 \\
DNI$_{t-9}$ & 3.06 \\
\bottomrule
\end{tabular}
\end{table}

The impact of the noise model is also evaluated during this step. \autoref{tab:noise_model} shows the validation error across all three train-validation data set splits. The architecture including the noise signal  performs consistently better than the case without the noise model, and often significantly better. More importantly, using the noise signal input eliminates the drop in performance when the model is trained with fewer features---with a final MAE of 74.34 \wmm. This confirms our initial hypothesis that the noise model accounts for both the dropped features and other unmeasured disturbances.

\begin{table}[]
    \centering
    \caption{Validation error across all 3 steps of model refinement. Bold numbers indicate the best (or nearly-best) errors for each step. Results with and without the noise model as well as with all features and the top 10 features are shown. Results indicate that the noise model eliminates the drop in performance when the model is trained with fewer features.}
    \label{tab:noise_model}
    \begin{tabular}{|l|l|lll|}
    \hline
    \multirow{2}{*}{Noise Model} & \multirow{2}{*}{Features} & \multicolumn{3}{l|}{Validation Error (\wmm)}                             \\ \cline{3-5}
                                 &                           & \multicolumn{1}{l|}{Step 1}           & \multicolumn{1}{l|}{Step 2}           & Step 3           \\ \hline
    \multirow{2}{*}{Not Included}       & All                       & \multicolumn{1}{l|}{\textbf{89.35}} & \multicolumn{1}{l|}{76.85}          & 75.20          \\ \cline{2-5}
                                 & Top 10                    & \multicolumn{1}{l|}{94.61}          & \multicolumn{1}{l|}{83.93}          & 77.36          \\ \hline
    \multirow{2}{*}{Included}        & All                       & \multicolumn{1}{l|}{91.07}          & \multicolumn{1}{l|}{\textbf{76.01}} & \textbf{74.95} \\ \cline{2-5}
                                 & Top 10                    & \multicolumn{1}{l|}{\textbf{88.96}} & \multicolumn{1}{l|}{\textbf{76.01}} & \textbf{74.34} \\ \hline
    \end{tabular}
\end{table}

\section{Discussion}

\begin{figure*}
    \centering
    \includegraphics[width=7in]{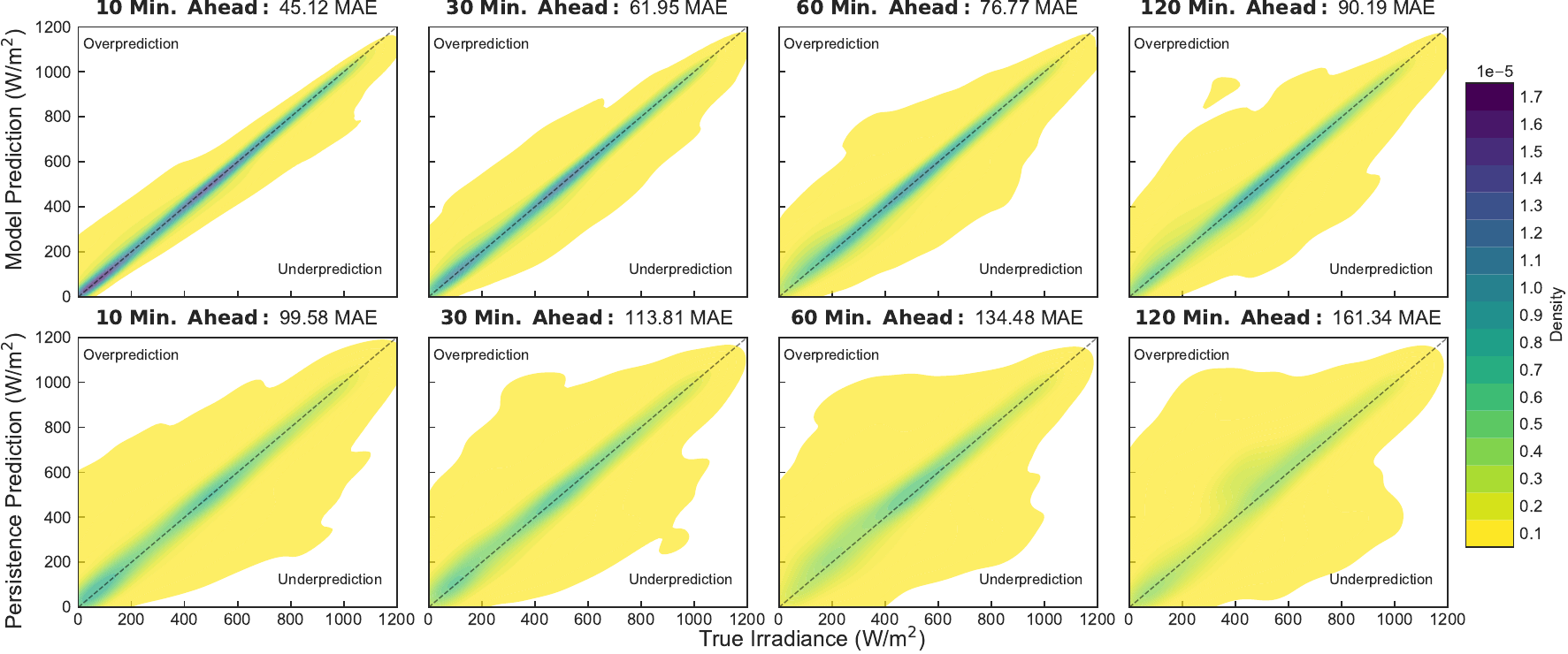}
    \caption{Mean Absolute Error (\wmm) distribution the model and POC. As expected, the error increases as time from the forecast (the time horizon) also increases. Density near the correct prediction is notably higher compared to the persistence of cloudiness model. MAE across all predicted intervals is 75.20 \wmm}
    \label{fig:residual_kde}
\end{figure*}

The final model achieves a MAE of 75.20 \wmm, while the POC model has an overall accuracy of 134.35 \wmm. The distribution of predictions and true values of the final model is shown in \autoref{fig:residual_kde} with the persistence model shown below for comparison. These graphs display the normalized density of predicted and true irradiances such that the integrated density on each image is equal to one. The dashed black line shows  a perfect forecast;  higher concentrations of the plotted prediction values near this line represent better models. The proposed model reduces the gap between true and predicted irradiance compared to the persistence model. However, the distributions become similar as the prediction horizon increases.

\begin{figure*}
    \centering
    \includegraphics[width=6.5in]{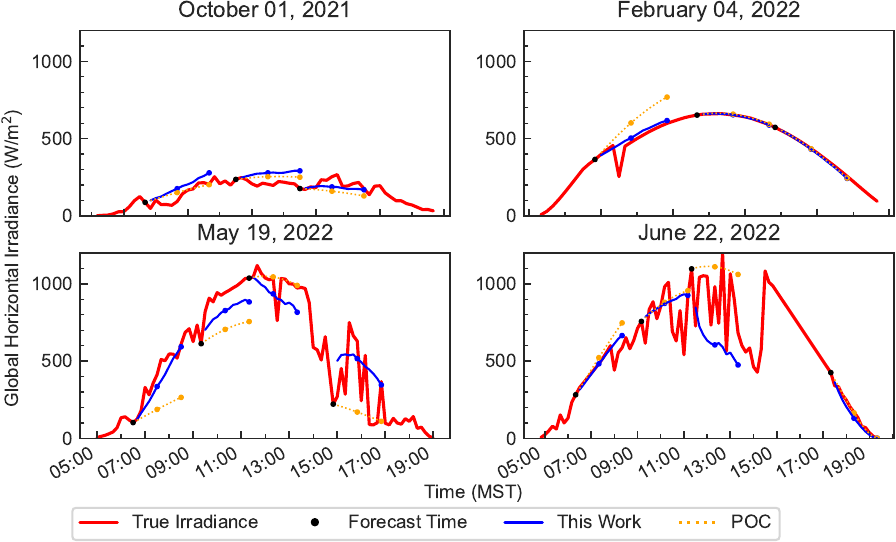}
    \caption{True irradiance as well as the predicted irradiance using the final model and POC across four days in the final validation set.}
    \label{fig:predictions}
\end{figure*}

Individual predictions are shown for four sample days in \autoref{fig:predictions}. Notably, POC and the proposed model perform similarly in environments with low irradiance (top left) or relatively constant environments (top right). On days with large ramping events in irradiance (bottom) the proposed model is better able to anticipate changes in regime, though sometimes these predictions are too aggressive. The results of this work are shown relative to recent literature values collected at the same location in \autoref{tab:rmse}. It is notable that the models proposed in the literature require sequences of images as inputs (and thus have high data transmission requirements)  while the model proposed in this work only uses the features extracted from the images, and is thus data-parsimonious.

% table
\begin{table*}
\centering
\caption{RMSE (W/m$^2$) of this work, POC, and reported errors from recent literature}
\label{tab:rmse}
\begin{tabular}{lllll}
\toprule
Forecast Horizon (Min.) & POC & This work & \cite{feng2022} & \cite{al-laham2020} \\
\midrule
10 & 264.56 & 97.90 & 71.30 & {-} \\
20 & 254.14 & 111.19 & 98.53 & {-} \\
30 & 258.72 & 119.86 & 109.33 & {-} \\
40 & 260.19 & 126.00 & 119.35 & {-} \\
50 & 264.17 & 131.69 & 127.49 & {-} \\
60 & 263.40 & 136.66 & 135.43 & 116.7 \\
70 & 263.86 & 139.82 & {-} & {-} \\
80 & 269.37 & 143.48 & {-} & {-} \\
90 & 268.16 & 145.36 & {-} & {-} \\
100 & 266.79 & 148.65 & {-} & {-} \\
110 & 265.10 & 151.49 & {-} & {-} \\
120 & 265.33 & 153.52 & {-} & 127.6 \\
\bottomrule
\end{tabular}
\end{table*}

Integrating the clear sky model and the persistence assumption explicitly into the forecast by changing the representation of future irradiance improves forecasts more than simply providing the clear sky irradiance. These representations simplify model training by isolating the unknown weather effects and not relying on the model to learn already-known long-term dynamics (i.e., de-trending). %This work extends the findings of \cite{gao2022} by predicting residual error of the persistence of cloudiness model rather than the residual of the clear sky model.
This work achieves a lower nMAP on the validation period as shown in \Cref{tab:nMAP} while using a dramatically simpler model. %Predicting irradiance relative to POC rather than clear sky irradiance by incorporating the conditions at the time of forecast and a persistence assumption~\cite{hyndman2021}.

% table
\begin{table}
\centering
\caption{nMAP (W/m$^2$) of POC, this work, and the best results from \cite{gao2022}}
\label{tab:nMAP}
\begin{tabular}{llllll}
\toprule
Forecast Horizon (Min.) & POC & This work &  \cite{gao2022}\\
\midrule
10 & 21.5 & 9.6 & {-} \\
20 & 22.8 & 11.9 & {-} \\
30 & 24.6 & 13.4 & {-} \\
40 & 26.2 & 14.7 & {-} \\
50 & 27.8 & 15.8 & {-} \\
60 & 29.1 & \textbf{16.8} & 17.4 \\
70 & 30.1 & 17.4 & {-} \\
80 & 31.6 & 18.0 & {-} \\
90 & 32.6 & 18.6 & {-} \\
100 & 33.4 & 19.2 & {-} \\
110 & 34.1 & 19.6 & {-} \\
120 & 34.9 & \textbf{20.0} & 20.9 \\
\bottomrule
\end{tabular}
\end{table}

\begin{figure}
    \centering
    \includegraphics[width=3.5in]{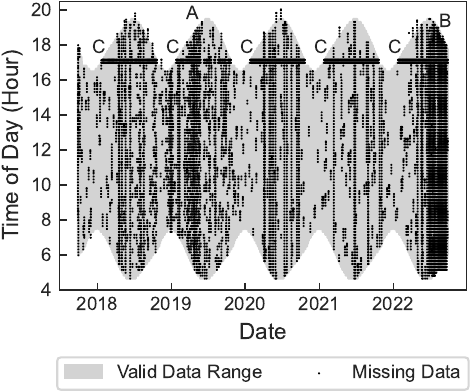}
    \caption{ASI-16 Images missing in the NREL SRRL BMS dataset. A and B were firmware issues noted in the documentation online while C notes a known bug where the camera does not take pictures near 17:00 MST (00:00 GMT)}
    \label{fig:missing_data}
\end{figure}

\autoref{fig:missing_data} shows missing data from the ASI-16 sky camera from outages or known firmware issues. Sky camera images are missing after about 17:00 MST in much of the dataset due to a firmware bug. The missing data may contribute to poor model performance, and may cause significant biases in model performance. Our approach was to only use segments of data without missing data-points. Other works using the NREL SRRL BMS such as \cite{gao2022,feng2022,al-laham2020} also selected only complete sequences. Future work could focus on devising more robust models that account for missing data from the sky camera.

\begin{figure}
    \centering
    \includegraphics[width=3.5in]{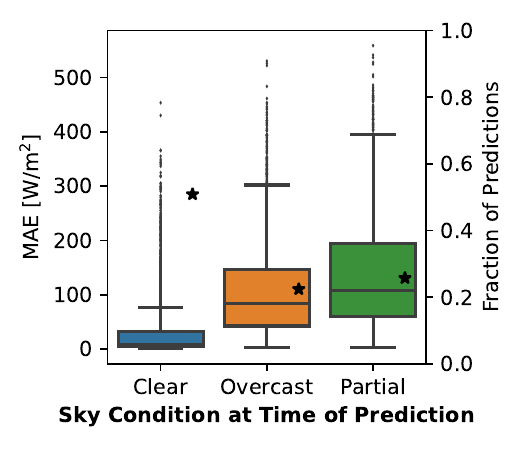}
    \caption{MAE of validation forecasts categorized by three distinct cloud cover conditions: Clear indicates CDOC Total Cloud Cover $<$ 20\%, Partially Cloudy indicates 20\% $<$ CDOC Total Cloud Cover $<$ 80\% and Overcast indicates CDOC Total Cloud Cover $<$ 80\%. In addition, the star marker for each cloud cover condition indicates the fraction of predictions within each category.}
    \label{fig:sky_conditions}
\end{figure}

\autoref{fig:sky_conditions} shows model prediction MAE boxplots for three distinct sky cover conditions at the time of forecast. The values were computed using the validation data. "Clear" indicates CDOC Total Cloud Cover $<$ 20\%, "Partially Cloudy" indicates 20\% $<$ CDOC Total Cloud Cover $<$ 80\% and "Overcast" indicates CDOC Total Cloud Cover $<$ 80\%. Intuitively, clear conditions are more easily predictable as they follow the persistence of cloudiness assumption. The MAE values for over half of the clear sky condition forecasts were small. However, the model may not anticipate changes in sky conditions and the accompanying changes in clear sky index. This is shown by the large volume of high error outliers in the clear sky category. Not unexpectedly, prediction accuracy drops for partially covered and overcast skies. We explain this by considering the naturally higher likelihood of a change in conditions over the forecast horizon from the time of prediction.

\section{Conclusions}

We developed a data-parsimonious machine learning model for near-term forecasting of solar irradiance. The model relies on a novel CNN-LSTM architecture that includes a noise signal input to account for random/unmeasured variables that influence irradiance.

A training process was proposed, that included probing data representation/feature engineering and feature selection using a sequence of training-validation data sets selected from a publicly available database. The focus was on predictions  up to two hours ahead at ten minute intervals, but we expect that other prediction frequencies and near-term horizons can be easily accommodated.

Our findings indicate that predicting the deviation of irradiance from a long-term baseline (e.g., the POC prediction) benefits from a de-trending effect and is thus more accurate than predicting the irradiance itself. We also found that including the noise model (even after eliminating a substantial number of features based on feature importance studies) leads to consistently higher prediction accuracy.

The validation data indicate that the proposed model outperforms other similar results available in the recent literature while requiring orders of magnitude less data as inputs.

\section*{Data Availability}
\label{sec:data}

The data used for this study were provided by \cite{stoffel1981}. Code used for data aggregation, modeling, and analysis is available at \url{https://github.com/joshuaeh/TabularSolarForecast}.

\section*{Declaration of Competing Interest}
\label{sec:declaration}

The authors declare that they have no conflict of interest.

\section*{Funding Source}
\label{sec:funding}

Financial support of this work was provided by the Robert A. Welch Foundation (F-1464), the Center for a Solar Powered Future (SPF2050)—an Industry-University Cooperative Research Center (IUCRC) funded by the National Science Foundation (EEC-2052814).

\section*{Acknowledgements}
\label{sec:acknowledgements}

The authors would like to thank Drs. Herie Soto, Dhruv Aurora, Hadi Jamali-Rad, Pierre Carrette and Ojas Shreikar for their feedback while working on this research.

\section*{CRediT Authorship Contribution Statement}
\label{sec:credit}

\textbf{Joshua Hammond}: Conceptualization, Methodology, Software, Validation, Formal Analysis, Data Curation, Writing - Original Draft, Visualization. \textbf{Ricardo Lara}: Methodology, Software, Validation, Writing - Review \& Editing. \textbf{Brian Korgel}: Conceptualization, Resources, Writing - Review \& Editing, Supervision, Funding Acquisition. \textbf{Michael Baldea}: Conceptualization, Methodology, Formal analysis, Writing - Review \& Editing, Supervision.

\appendix

\section{All Features Studied}
\label{sec:apendix-parameters}
The features used in this study are organized by source, with units shown in brackets when applicable:\footnote{More information on the NREL SRRL BMS measurements is available at \url{https://midcdmz.nrel.gov/apps/html.pl?site=BMS;page=instruments}}

\subsection*{BMS Meteorological Station Features:}
\begin{itemize}
    \setlength\itemsep{-0.2 ex}
    \item 315nm Photometer [nA]
    \item 400nm Photometer [uA]
    \item 500nm Photometer [uA]
    \item 675nm Photometer [uA]
    \item 870nm Photometer [uA]
    \item 940nm Photometer [uA]
    \item 1020nm Photometer [uA]
    \item Snow Depth [cm]
    \item Precipitation [mm]
    \item Precipitation (Accumulated) [mm]
    \item Station Pressure [mBar]
    \item Tower Dry Bulb Temperature [deg C]
    \item Tower Relative Humidity [\%]
    \item Snow Depth Quality [\%]
    \item Station Dry Bulb Temp [deg C]
    \item Station Relative Humidity [\%]
    \item Vertical Wind Shear [1/s]
    \item Average Wind Speed at 22ft [m/s]
    \item Average Wind Direction at 22ft [deg from N]
    \item Peak Wind Speed at 22ft [m/s]
    \item Albedo (CM3)
    \item Albedo (LI-200)
    \item Albedo Quantum (LI-190)
    \item Broadband Turbidity
    \item Sea-Level Pressure (Est) [mBar]
    \item Tower Dew Point Temp [deg C]
    \item Tower Wet Bulb Temp [deg C]
    \item Tower Wind Chill Temp [deg C]
    \item Airmass
    \item GHI [W/m$^2$]
    \item DNI [W/m$^2$]
    \item DHI [W/m$^2$]
\end{itemize}

\subsection*{Sky Camera Features:}
\begin{itemize}
    \setlength\itemsep{-0.2 ex}
    \item Blue-Red/Blue-Green Total Cloud Cover [\%]
    \item Cloud Detection and Opacity Correction Total Cloud Cover [\%]
    \item Cloud Detection and Opacity Correction Thick Cloud Cover [\%]
    \item Cloud Detection and Opacity Correction Thin Cloud Cover [\%]
    \item Haze Correction Value
    \item Blue/Red minimum
    \item Blue/Red median
    \item Blue/Red maximum
    \item Apparent Solar Zenith Angle [deg]
    \item Apparent Solar Azimuth Angle [deg]
    \item Flag: Sun not visible
    \item Flag: Sun on clear sky
    \item Flag: Parts of sun covered
    \item Flag: Sun behind clouds, bright dot visible
    \item Flag: Sun outside view
    \item Flag: No evaluation
\end{itemize}

\subsection*{Clear Sky Model Features:}
\begin{itemize}
    \setlength\itemsep{-0.2 ex}
    \item Clear Sky GHI [W/m$^2$]
    \item Clear Sky DNI [W/m$^2$]
    \item Clear Sky DHI [W/m$^2$]
    \item Solar Eclipse Shading
    \item Zenith Angle [deg]
    \item Solar Elevation Angle [deg]
    \item Solar Azimuth Angle [deg]
\end{itemize}

\subsection*{Engineered Features:}
\begin{itemize}
    \setlength\itemsep{-0.2 ex}
    \item Time from sunrise [Days]
    \item Time to solar noon [Days]
    \item Time to sunset [Days]
    \item Cosine time from sunrise [Days]
    \item Sine time from sunrise [Days]
    \item Cosine time to solar noon [Days]
    \item Sine time to solar noon [Days]
    \item Cosine time to sunset [Days]
    \item Sine time to sunset [Days]
    \item Flag: Day
    \item Flag: Before solar noon
    \item Cosine zenith angle
    \item Cosine normal irradiance
    \item Wind North-South Speed [m/s]
    \item Wind East-West Speed [m/s]
    \item Sun North-South Position
    \item Sun East-West Position
    \item Time of Day [Days]
    \item Time of Year [Years]
    \item Sine Time of Year [Years]
    \item Cosine Time of Year [Years]
    \item Sine Time of Day [Days]
    \item Cosine Time of Year [Days]
    \item Clear Sky Index GHI
    \item Clear Sky Index DNI
    \item Clear Sky Index DHI
    \item GHI$_{t-1}$ [W/m$^2$]
    \item DNI$_{t-1}$ [W/m$^2$]
    \item DHI$_{t-1}$ [W/m$^2$]
    \item GHI$_{t-2}$ [W/m$^2$]
    \item DNI$_{t-2}$ [W/m$^2$]
    \item DHI$_{t-2}$ [W/m$^2$]
    \item GHI$_{t-3}$ [W/m$^2$]
    \item DNI$_{t-3}$ [W/m$^2$]
    \item DHI$_{t-3}$ [W/m$^2$]
    \item GHI$_{t-4}$ [W/m$^2$]
    \item DNI$_{t-4}$ [W/m$^2$]
    \item DHI$_{t-4}$ [W/m$^2$]
    \item GHI$_{t-5}$ [W/m$^2$]
    \item DNI$_{t-5}$ [W/m$^2$]
    \item DHI$_{t-5}$ [W/m$^2$]
    \item GHI$_{t-6}$ [W/m$^2$]
    \item DNI$_{t-6}$ [W/m$^2$]
    \item DHI$_{t-6}$ [W/m$^2$]
    \item GHI$_{t-7}$ [W/m$^2$]
    \item DNI$_{t-7}$ [W/m$^2$]
    \item DHI$_{t-7}$ [W/m$^2$]
    \item GHI$_{t-8}$ [W/m$^2$]
    \item DNI$_{t-8}$ [W/m$^2$]
    \item DHI$_{t-8}$ [W/m$^2$]
    \item GHI$_{t-9}$ [W/m$^2$]
    \item DNI$_{t-9}$ [W/m$^2$]
    \item DHI$_{t-9}$ [W/m$^2$]
    \item Clear Sky Deviation GHI$_{t}$ [W/m$^2$]
    \item Clear Sky Deviation DNI$_{t}$ [W/m$^2$]
    \item Clear Sky Deviation DHI$_{t}$ [W/m$^2$]
    \item Mean Clear Sky Deviation GHI$_{t-10:t}$ [W/m$^2$]
    \item Mean Clear Sky Deviation DNI$_{t-10:t}$ [W/m$^2$]
    \item Mean Clear Sky Deviation DHI$_{t-10:t}$ [W/m$^2$]
    \item Median Clear Sky Deviation GHI$_{t-10:t}$ [W/m$^2$]
    \item Median Clear Sky Deviation DNI$_{t-10:t}$ [W/m$^2$]
    \item Median Clear Sky Deviation DHI$_{t-10:t}$ [W/m$^2$]
    \item Clear Sky Index Standard Deviation GHI$_{t-10:t}$ [W/m$^2$]
    \item Clear Sky Index Standard Deviation DNI$_{t-10:t}$ [W/m$^2$]
    \item Clear Sky Index Standard Deviation DHI$_{t-10:t}$ [W/m$^2$]
    \item Mean Clear Sky Deviation GHI$_{t-60:t}$ [W/m$^2$]
    \item Mean Clear Sky Deviation DNI$_{t-60:t}$ [W/m$^2$]
    \item Mean Clear Sky Deviation DHI$_{t-60:t}$ [W/m$^2$]
    \item Median Clear Sky Deviation GHI$_{t-60:t}$ [W/m$^2$]
    \item Median Clear Sky Deviation DNI$_{t-60:t}$ [W/m$^2$]
    \item Median Clear Sky Deviation DHI$_{t-60:t}$ [W/m$^2$]
    \item Clear Sky Index Standard Deviation GHI$_{t-60:t}$ [W/m$^2$]
    \item Clear Sky Index Standard Deviation DNI$_{t-60:t}$ [W/m$^2$]
    \item Clear Sky Index Standard Deviation DHI$_{t-60:t}$ [W/m$^2$]
\end{itemize}

% If you have bibdatabase file and want bibtex to generate the
% bibitems, please use
%
% \bibliographystyle{elsarticle-num}
% \bibliographystyle{unsrtnat}
% \bibliography{refs}
% \begin{thebibliography}{}
%     \input{main.bbl}
% \end{thebibliography}
% \input{main.bbl}

\end{document}